\pdfoutput=1

\documentclass[11pt]{article}

\usepackage{acl}

\usepackage{times}
\usepackage{latexsym}
\usepackage{float}
\usepackage[T1]{fontenc}

\usepackage[utf8]{inputenc}

\usepackage{microtype}

\usepackage{inconsolata}

\usepackage{graphicx}
\usepackage{multirow}
\title{Towards Typologically Aware Rescoring to Mitigate Unfaithfulness in Lower-Resource Languages}

\author{{\bf Tsan Tsai Chan\thanks{Equal contributors.}}, {\bf Xin Tong\footnotemark[\value{footnote}]}, {\bf Thi Thu Uyen Hoang\footnotemark[\value{footnote}]} \\ {\bf Barbare Tepnadze}, {\bf Wojciech Stempniak} \\
Saarland University, 66123 Saarbrücken, Germany\\ \texttt{\{tsch00001, xito00001, thho00003, bate00001, wost00001\}}\\\texttt{@stud.uni-saarland.de}}

\begin{document}
\maketitle
\begin{abstract}
Multilingual large language models (LLMs) are known to more frequently generate non-faithful output in resource-constrained languages \citep[cf.][]{guerreiro2023hallucinations}, potentially because these typologically diverse languages are underrepresented in their training data. To mitigate unfaithfulness in such settings, we propose using computationally light auxiliary models to rescore the outputs of larger architectures. As proof of the feasibility of such an approach, we show that monolingual 4-layer BERT models pretrained from scratch on less than 700 MB of data without fine-tuning are able to identify faithful summaries with a mean accuracy of 88.33\% in three genetically unrelated languages that differ in their morphological complexity --- Vietnamese, Polish and Georgian. The same hyperparameter combination moreover generalises well to three other tasks, suggesting applications for rescoring beyond improving faithfulness. In order to inform typologically aware model selection, we also investigate how morphological complexity interacts with regularisation, model depth and training objectives, ultimately demonstrating that morphologically complex languages are more likely to benefit from dropout, while across languages downstream performance is enhanced most by shallow architectures as well as training using the standard BERT objectives.
\end{abstract}

\section{Introduction}

Large multilingual foundation models often underperform in lower-resource languages, most notably suffering significantly higher hallucination rates than English \citep{berbatova2023evaluating, leong2023bhasa}. This study takes a first step towards exploring the potential of rescoring as a computationally undemanding means of mitigating faithfulness-related hallucination in generative large language models (LLMs). More concretely, that would involve a smaller model reranking the output of a larger one to optimise a faithfulness objective, ultimately aiming to make the system's outputs more faithful. Rescoring has been productively applied in many fields to enhance the output of large architectures, with speech processing a prime example \citep[e.g.][]{khedar2024automatic}. It has considerable potential for lower-resource languages, as the small classification models required do not need to be trained on large amounts of data.

In the following, we show that monolingual BERT models pretrained from scratch using conventional objective functions without fine-tuning on less than 700 MB of data per language can discern faithful summaries in a binary context at a level between 26.25 and 50\% above chance. We verify this for three typologically distinct languages --- Vietnamese (morphologically simple), Polish and Georgian (morphologically complex). Crucially, we also demonstrate that these models generalise well to three other tasks, which are not especially closely related. Although they are not actually used to rescore LLM outputs here, this is compelling evidence that such encoder-only models could feasibly be used for multi-purpose rescoring.

To shed more light on what adjustments might be needed to adapt such models to languages with differing degrees of morphological complexity, we additionally study how morphological complexity interacts with regularisation, model depth and training objectives, 
finding that: 
\begin{itemize}
\setlength{\itemsep}{0pt}
\item Shallower models produce more robust performance across tasks regardless of language, \textit{pace} \citet{otmakhova2022cross}.
\item Dropout may benefit morphologically complex languages, while appreciably harming performance for Vietnamese.
\item The standard BERT training objectives \citep{devlin2019bert} are robust across languages.
\end{itemize}
These findings at once buttress the case for auxiliary models as a computationally undemanding means of improving LLM output in lower-resource languages and motivate further investigation into how typological differences between languages affect downstream performance.

\section{Related Work}
\subsection{Hallucination in lower-resource languages}
In the context of generative LLMs, hallucination describes a variegated array of phenomena involving the output either not being factual or otherwise not adhering to the user's instructions \citep{huang2023survey}. Hallucinated output has serious safety and ethical implications, as it could for instance contribute directly to misinformation \citep{huang2023survey}. For tractability, this study focusses only on faithfulness-related hallucination, in particular scenarios where LLMs produce inaccurate summaries of longer texts \citep[cf.][]{qiu2023detecting}.

Much work has indicated that insufficient good-quality data is one of the main factors behind unfaithfulness and other forms of hallucination \citep{huang2023survey}. Unfortunately, the low-data regimes that many lower-resource languages instantiate effectively rule out data-hungry methods such as retrieval-augmented generation (RAG) that have been used to mitigate hallucination with some success \citep[cf.][]{kirchenbauer2024hallucination, tonmoy2024comprehensive}. In this study, we hence propose rescoring as a computationally light alternative to deal with unfaithful generation.

\subsection{Rescoring}
\label{sec:rescoring}
Rescoring, also known as reranking, is a set of methods especially common in machine translation and speech recognition, where an auxiliary model, a BERT variant for example, is used to refine the outputs of a larger, mostly generative, foundation model \citep{khedar2024automatic}. Because such auxiliary models typically do not require vast amounts of data to train, they have been used for lower-resource languages for many years now, especially in situations implicating information retrieval \citep[e.g.][]{lee2014graph, litschko2022parameter}. 

Discriminative reranking in particular has been productively applied to various use cases \citep{shen2004discriminative, mizumoto2016discriminative}. It involves the larger generative model producing \textit{N} best candidate outputs via beam search for instance, whereupon the smaller auxiliary model rescores these to optimise a ranking objective and lead to more robust decisions. The auxiliary model could either be coupled to the foundation model for inference or used to fine-tune it. However, to the best of our knowledge, the potential of rescoring has thus far not been exploited to improve LLM faithfulness. It is moreover unclear to what extent the hyperparameters of auxiliary models need to be adapted to the typological features of each low-resource language for optimal performance.

\subsection{Morphology and hyperparameter choice}
\label{sec:morphology_hyperparameters}
The last point above is especially crucial as lower-resource languages have hugely diverse typological traits, forming a class only by virtue of not being English \citep{jin2024better}. Despite the widespread application of auxiliary models to these languages, not much research exists to our best knowledge on whether or how to optimise the models' hyperparameters in a manner sensitive to their morphological characteristics. \citet{otmakhova2022cross} is a partial exception. For a more nuanced perspective, we approach the issue from the following three angles:

\textbf{Regularisation}. It is widely reported that morphologically simple languages tend to have a lower type-to-token ratio, which correlates very strongly with lower perplexity \citep[e.g.][]{mielke2019kind, levshina2022frequency}. We could therefore theorise that in low-data regimes, models trained on morphologically simple languages would be more exposed to overfitting than those trained on morphologically complex ones and would benefit from stronger regularisation. This assumption finds support in \citet{otmakhova2022cross}'s observations on model depth, which the next paragraph turns to.

\textbf{Model depth}. Work on this front is more limited, but \citet{otmakhova2022cross}'s study shows that complex morphology is learnt in the upper layers of monolingual BERT-Base models. This leads them to conjecture that morphologically complex languages require deeper models than morphologically simpler ones to learn properly. Conversely, they also report that overly deep models trained on morphologically simpler languages cause a drop in performance, potentially indicating that shallow architectures would be best for these languages. 

\textbf{Training objectives}. To train the original BERT, masked language modelling (MLM) and next-sentence prediction (NSP) were used in conjunction \citep{devlin2019bert}. These are, however, not the only options, with one notable alternative to the latter objective being sentence-order prediction (SOP) \citep{lan2019albert}. While its designers claim SOP to potentially be more helpful for tasks relating to discourse coherence, we are not aware of work that has rigorously tested if the purported advantages of alternative objective functions hold across languages. Given that morphologically complex languages are associated with freer word order \citep[e.g.][]{nijs2025word}, we would expect NSP to be less effective for these languages especially in low-resource settings, as their flexible word order would lead to more possible variations of any given sentence and correspondingly higher perplexity. SOP, on the other hand, deals with higher-level discourse features less directly connected to word order, and could thus be justifiably hypothesised to better facilitate the modelling of morphologically complex languages.

\section{Experiment}
\subsection{Choice of languages}
\label{sec:langs}
As mentioned above, this study encompasses one morphologically simple language, Vietnamese, and two morphologically highly complex ones, Polish and Georgian. Vietnamese is classed as an isolating language as its verbs and nouns do not take inflectional affixes that mark for instance person, number or gender \citep{comrie1989language}. On the other hand, Polish and Georgian are both very affix-rich but typologically distinct from each other --- Polish is fusional, meaning many of its affixes encapsulate more than one meaning, while Georgian is agglutinative, implying its affixes have a neater one-to-one correspondence with grammatical meaning \citep{comrie1989language}. Interestingly, agglutinative languages are reportedly harder than fusional ones to model \citep{gerz2018relation} despite both classes having complex morphology in common.

\subsection{Hypotheses}
\label{sec:hypotheses}
Based on the work cited in \ref{sec:morphology_hyperparameters}, we put forward three hypotheses relating morphological complexity to hyperparameter choice, which the rest of this paper will test:

\textbf{1}. Morphologically simple languages (Vietnamese in our sample) are more prone to causing overfitting and would hence benefit from regularisation more than morphologically complex languages (Polish and Georgian).

\textbf{2}. Morphologically complex languages produce the best performance with deeper models, whereas morphologically simple ones require shallower architectures for equivalent performance.

\textbf{3}. Sentence-order prediction (SOP) aids the learning of morphologically complex languages more strongly than next-sentence prediction (NSP).

\subsection{Datasets}
Every model variant is trained on only one language, each language's dataset comprising approx. 227,281,794 characters' worth of Wikipedia articles and around twice as many characters (466,844,003) from news corpora. These proportions were enforced by soft caps, which were in turn based on character counts for the Georgian Wikipedia dump and Polish news corpus respectively. No documents were truncated in applying these caps, and each dataset was eventually tokenised using language-specific subword and sentence-level tokenisers. The capped Wikipedia and news data for each language was less than 700 MB in size before subsequent processing to accommodate the objective functions. For more details on the datasets and tokenisers used in this study, see Appendix \ref{sec:appendix}. 

\subsection{Hyperparameters varied}
We vary those hyperparameters that directly relate to our three hypotheses in \ref{sec:hypotheses}, holding learning rate decay and weight decay constant:

\textbf{Regularisation}. We apply two levels of \textbf{dropout} to the feedforward layers, 0 and 0.1. We also vary \textbf{batch size} between our default of 128 and the alternative setting of 256, but only for our 4-layer MLM+NSP models owing to limited compute.

\textbf{Model depth}. Our models have depths of 4, 8 and 12 layers. The number of attention heads per layer is kept constant at 8 for all model variants, such that the 4 and 8-layer ones approximate the architecture of BERT Small and BERT Medium respectively \citep{turc2019}.

\textbf{Training objectives}. We experiment with two combinations of these, masked language modelling and next-sentence prediction (MLM+NSP) and masked language modelling and sentence-order prediction (MLM+SOP). We did not combine all three objectives due to computational limitations.

\subsection{Evaluation tasks}
\label{sec:tasks}
Prompts for most tasks were originally in English, either already available or crafted from scratch. They were then automatically translated and corrected by native speakers of each language, including the last three authors. All tasks were set up as binary classification.

\textbf{Faithfulness detection}. This task provided the model a paragraph of moderate length in the format \texttt{<PARAGRAPH> <'In short,'>} and required it to select the more faithful summary from two options. We used a total of 80 prompts based on 20 paragraphs, which were extracts from articles on Wikinews \citep{wikinews}. We used ChatGPT to generate faithful summaries of these paragraphs, which we then manually modified to produce four summaries in total:

\textbf{1}. \textit{Faithful (non-adversarial)}. Preserves content, vocabulary, sentence structure and order of points.

\textbf{2}. \textit{Faithful (adversarial)}. Preserves original content but changes phrasing and order of points.

\textbf{3}. \textit{Unfaithful (non-adversarial)}. Preserves original structure, with some words replaced with irrelevant and factually incorrect ones.

\textbf{4}. \textit{Unfaithful (adversarial)}. Preserves original structure, with some words replaced with contextually relevant but factually incorrect ones.

Table \ref{tab:faithfulness_example} illustrates how faithful and unfaithful summaries were derived from one original paragraph. For each prompt, the model was supplied the paragraph along with one faithful summary and an unfaithful one. Our aim was to test the models' ability to pick out faithful summaries regardless of sentence structure, the sequence of points or choice of words, somewhat analogously to the XL-Sum benchmark \citep{hasan2021xlsum}.

\begin{table*}
  \centering
  \small 
  \resizebox{\textwidth}{!}{%
  \begin{tabular}{p{0.25\textwidth} p{0.16\textwidth} p{0.18\textwidth} p{0.16\textwidth} p{0.16\textwidth}}
    \hline
    \textbf{Original paragraph} & \textbf{Correct option (non-adversarial)} & \textbf{Correct option (adversarial)} & \textbf{Incorrect option (non-adversarial)} & \textbf{Incorrect option (adversarial)}  \\
    \hline
    ... oxygen is essential for advanced living organisms. Some species of microorganisms do not require oxygen for metabolism, [...] such as methanogens which rely on carbon dioxide while releasing methane. In short, & … oxygen is crucial for complex organisms, though some microorganisms, like methanogens, thrive without oxygen, using carbon dioxide instead. & … complex organisms cannot do without oxygen, though some microorganisms, like methanogens, thrive in environments without it, instead using carbon dioxide. & …  \underline{money} is crucial for complex organisms, though some \underline{recluses}, like methanogens, thrive without \underline{money}, using carbon \underline{offsets} instead. & … \underline{water} is crucial for complex organisms, though some \underline{viruses}, like methanogens, thrive without \underline{water}, using carbon \underline{monoxide} instead. \\
    \hline
  \end{tabular}
  }
  \caption{Abridged example (in English) of correct/ incorrect and adversarial/ non-adversarial options in the faithfulness detection task. Words unfaithful to the original context are underlined. One correct option is always paired with one incorrect option, making for a total of four possible prompts for each original paragraph.}
  \label{tab:faithfulness_example}
\end{table*}

\textbf{Other tasks}. In order to test the generalisability of our models to other NLP-related applications, we evaluate our models on three other tasks that were chosen for their relevance to natural language understanding and because they did not appear directly connected to either of our training objective combinations. Like faithfulness detection, all these tasks were set up as binary classification:
\begin{itemize}
    \setlength{\itemsep}{0pt}
    \item \textbf{Real-world knowledge} (90 prompts). Given the definition of an entity, the model needs to choose the entity being referred to. Prompts were adapted from the RELPRON dataset \citep{rimell2016relpron}, with correct answers mostly from the original prompts and incorrect ones being the answers to other prompts. We modified our Polish and Georgian prompts to make sure that each pair of options had the same number of word stems.
    \item \textbf{Textual entailment} (30 prompts). Given a context, the model needs to select the pragmatically correct continuation. Three types of entailment were included --- causal, temporal and contrastive.
    \item \textbf{Winograd Schema Challenge} (36 prompts). Given a context implicating two people, the model needs to infer via pronoun resolution which person a subsequent sentence refers to. To adapt the prompts for MLM in Polish and Georgian, we masked only stems and left affixes unmasked, separated from the \texttt{<MASK>} token by a whitespace token.
\end{itemize}

Table \ref{tab:tasks_comparison} in Appendix \ref{sec:appendix} contains examples of our prompts, whose modest numbers here were a consequence of more comprehensive benchmarks not being available in all the languages covered in this study and the infeasibility of manually translating much larger quantities of prompts.

\section{Results}
The results of our experiments largely contradict the three hypotheses laid out in \ref{sec:hypotheses}, showing that:

\textbf{1}. Increased regularisation is on average more beneficial to accuracy in our two morphologically complex languages than Vietnamese.

\textbf{2}. Shallow architectures generalise the best across our morphologically distinct languages as opposed to only being suited for learning morphologically simple languages, and

\textbf{3}. MLM+NSP tends to be more helpful for performance than MLM+SOP, although the disparity is less pronounced in Georgian.

Below, we examine these results in greater detail, looking at faithfulness detection before covering our models' generalisation to the other tasks.
\subsection{Faithfulness detection}
We first present the hyperparameter combinations that produced the highest accuracy scores for this task. After that, we turn to the hypotheses in \ref{sec:hypotheses}.

\textbf{Most typologically robust setting}. The best compromise among our three languages is MLM+NSP | 0.1 dropout | \underline{4 layers} | batch size 128, with accuracy in Vietnamese at 93.75\%, Polish at 100\% and Georgian at 71.25\% (mean 88.33\%). This shallow configuration outperformed the 8 and 12-layer models by a mean of 23.02\%, a strong indication that lightweight set-ups can excel at faithfulness detection. Henceforth, we assume batch size to be 128 unless stated otherwise.\footnote{Where appropriate, we specify our model variants by stating their hyperparameter settings in this format: \texttt{<training objectives> | <dropout setting> | <no. of layers>}.} 

\textbf{Best settings by language}. The following language-specific hyperparameters attained the best accuracy rates, again featuring 4-layer models:

\begin{itemize}
    \setlength{\itemsep}{0pt}
    \item \textit{Vietnamese} (95\%): MLM+NSP | 0 dropout | \underline{4 layers} | batch size 128
    \item \textit{Polish} (100\%): \underline{All MLM+NSP models}, MLM+SOP | 0 dropout | 12 layers and MLM+SOP | 0.1 dropout | 8 layers (batch sizes 128, 256 for all)
    \item \textit{Georgian} (76.25\%): MLM+NSP | 0.1 dropout | \underline{4 layers} | batch size 256
\end{itemize}
We now address our hypotheses in turn, using MLM+NSP | batch size 128 as our default.

\textbf{Regularisation}. We see some indication that increasing regularisation by adding dropout and decreasing batch size consistently improves faithfulness detection in Vietnamese, in line with our hypothesis. However, there is no clear-cut evidence suggesting regularisation uniformly hurts performance in our two morphologically complex languages. Adding 0.1 dropout to the MLM+NSP models improved mean accuracy across model depths in Vietnamese by +1.67\% (difference w/ dropout -1.25 to 8.75\%) and in Georgian by 2.92\% (diff. -7.5 to 8.75\%). It consistently made no difference to the 100\% accuracy rates of our Polish models (see Table \ref{tab:mlm_nsp}). For our 4-layer MLM+NSP models, decreasing batch size from 256 to 128 led to a more marked 5.63\% improvement for Vietnamese, again no difference for Polish and a significant 7.5\% deterioration for Georgian.

\textbf{Model depth}. While increasing depth for the MLM+NSP models both with and without dropout massively degrades accuracy in Vietnamese (-32.5\% from 4 layers to 12 w/o dropout), our Georgian models also suffer monotonically deteriorating accuracy (-17.9\% from 4 to 12 layers). The latter is unexpected by our hypothesis. Yet again, we observe no difference for Polish.

\textbf{Training objectives}. Across depths and dropout levels, our MLM+SOP models fare far worse than our MLM+NSP ones for all three languages. This goes against our theory that MLM+SOP would benefit morphologically complex languages more than MLM+NSP would do, but the less negative results for Polish (-15\%) and Georgian (-17.92\%) compared to Vietnamese (-26.49\%) suggests a weaker version of the hypothesis may still hold some truth.

\subsection{Mean performance across tasks}
\label{sec:mean_perf}
We now report our results averaged across models, languages and all four of the tasks in \ref{sec:tasks}. Task-by-task accuracy rates are presented in full for each language and model in Appendix \ref{sec:appendix_results}.

\textbf{Most typologically robust setting}. The best compromise across tasks is MLM+NSP | 0.1 dropout | 8 layers | batch size 128, with accuracy in Vietnamese at 61.66\%, Polish at 71.23\% and Georgian at 61.51\% (mean 64.80\%). However, the MLM+NSP | 0.1 dropout | \underline{4 layers} setting that performed best across languages for faithfulness detection is a close contender (64.77\%), along with MLM+NSP | 0 dropout | \underline{4 layers} (64.70\%, batch size 128 for both), again convincingly highlighting the generalisability of shallow models.

\textbf{Best settings by language}. These mostly overlap with the optimal faithfulness detection settings:

\begin{itemize}
    \setlength{\itemsep}{0pt}
    \item \textit{Vietnamese} (68.23\%): MLM+NSP | 0 dropout | \underline{4 layers} | batch size 128
    \item \textit{Polish} (75.12\%): MLM+NSP | 0 dropout | \underline{4 layers} | batch size 256; also MLM+NSP | 0.1 dropout | 12 layers | batch size 128 (74.92\%)
    \item \textit{Georgian} (62.09\%): MLM+NSP | 0.1 dropout | \underline{4 layers} | batch size 256
\end{itemize}

\textbf{Regularisation}. On balance, dropout harms accuracy in the MLM+NSP models in Vietnamese (-1.38\%), while improving it for Polish (+3.70\%) and Georgian (+1.20\%). This is at odds with our prediction that regularisation would benefit morphologically simpler languages, while affecting complex ones adversely. Controlling for depth and switching to the MLM+SOP setting did not produce unambiguous evidence supporting this prediction either. Similarly, the stronger regularisation brought about by decreasing batch size at 4 layers in the MLM+NSP setting actually \textit{degrades} accuracy more significantly in Vietnamese (-9.64\%) than in Polish (-1.03\%) or Georgian (+5.18\%).

\textbf{Model depth}. The cross-task figures conform better to our prediction that deeper models would undermine performance for Vietnamese, with mean accuracy in the MLM+NSP setting for that language severely degrading with increasing depth (-13.80\% w/o dropout from 4 layers to 12). Even so, there is no straightforward indication that deeper models benefit Polish or Georgian across the board. Again moving from 4 to 12 layers, we see a slight deterioration in Polish (-1.90\%) and only a modest improvement in Georgian (+1.74\%), which are slightly tempered by applying 0.1 dropout (-0.04\% and -1.44\% respectively). Switching to the MLM+SOP setting, however, we see improvements brought about by increased depth for Vietnamese and Polish of 3.46\% and 6.58\% respectively, while for Georgian there is a \textit{decrease} of 5.46\%, implying no obvious relationship between morphological complexity and model depth.

\textbf{Training objectives}. Across model depths and dropout settings, training models using MLM+SOP instead of MLM+NSP causes performance for Vietnamese and Polish to deteriorate by 6.06\% and 8.06\% respectively. The degradation is less extreme for Georgian, at 3.81\%. Controlling for dropout and model depth did not yield any evidence in favour of our prediction that MLM+SOP would be more suitable than MLM+NSP for morphologically complex languages, although all in all Georgian appears to take better to MLM+SOP than the two other languages.

In sum, our results show that the best hyperparameter settings for faithfulness detection largely transfer over to our other tasks, confirming the potential of shallow models for multi-purpose rescoring. While they do not lend credence to the hypotheses in \ref{sec:hypotheses}, these results point to morphology directly influencing optimal hyperparameters. 

\section{Discussion and Recommendations}
Before concluding, we briefly explore why our first two hypotheses in \ref{sec:hypotheses} do not hold and touch on how a rescoring system could be implemented.

\textbf{Regularisation}. Overturning our original hypothesis, our morphologically complex languages, Georgian in particular, benefited more than Vietnamese from regularisation --- recall in fact that the best Georgian model had 0.1 dropout (\ref{sec:mean_perf}). We speculate that the high type-to-token ratios associated with complex morphological systems --- especially agglutinative paradigms like Georgian's (but see Table \ref{tab:counts}) --- could encourage overfitting, but this, as well as the interaction between dropout and batch size, warrants more targeted investigation.

\textbf{Model depth}. \textit{Pace} \citet{otmakhova2022cross}, we found no clear interaction between model depth and morphological complexity. 
We speculate that our controlling for character count in each language's dataset may have had a similar effect to 
the set-up in \citet{arnett2025language}, who report no significant effect of morphology on perplexity after scaling their datasets according to each language's memory requirements. Regardless, all three of our languages favouring shallower architectures bodes well for our faithfulness rescoring approach, which aims to minimise computational overhead.

\textbf{Possible implementation}.
Assuming the accuracy rates reported here would also hold in the real world, we recommend evaluating the discriminative reranking (\ref{sec:rescoring}) capabilities of our language-specific and generic best architectures as a first step, training them on more diverse genres. Framing the task as binary classification, the \textit{N} best candidate summaries generated by an LLM in a lower-resource language would be compared pairwise against each other, with the auxiliary model ideally selecting the most faithful one via rescoring.

\section{Conclusion}
In the foregoing, we demonstrated the potential of rescoring using lightweight auxiliary models to improve LLM faithfulness in lower-resource languages. We explored how various hyperparameter combinations affect performance and showed that shallower BERT models generalise robustly across three typologically distinct languages and tasks, with regularisation benefiting morphologically complex languages more than expected. We hope that these findings will facilitate future work on typologically attuned rescoring pipelines.

\section*{Limitations}
The most obvious limitation of this study is its not having applied the auxiliary models discussed to any actual faithfulness rescoring task involving LLM output. In addition, we have limited ourselves to only BERT and three languages with alphabet-based scripts. Future work should explore a wider variety of auxiliary model architectures and a more targeted sample of morphologically distinct languages. In addition, our 256-token context width constrained us to relatively short texts for the faithfulness detection task, on average 150 to 250 tokens in length in the original English. Simulating more realistic settings may necessitate the use of longer texts and larger context windows, as well as multi-class classification, which could be relevant to non-English languages with larger amounts of data available.

\section*{Acknowledgements}
We would like to thank Michael Sullivan for his guidance and valuable advice as well as Paweł Jasiński, Wojciech Kopański and Bartek Pogodziński for meticulously proofreading the Polish prompts. Needless to say, we assume sole responsibility for any shortcomings and errors that may remain.

\bibliography{main}

See overleaf for appendices.
\clearpage
\appendix

\section{Datasets and Tokenisers Used}
\label{sec:appendix}


All Wikipedia data originates from the 1 November 2023 page dump by the Wikimedia Foundation on Hugging Face \citep{wikidump}. The news corpora we used were BKAI for Vietnamese \citep{duc2024towards}, polish-news for Polish \citep{pl-news},  and for Georgian a subset of the webcrawl dataset by \citet{RichNachos_Georgian_Corpus}, with only documents belonging to 32 news-related domains retained.

We employed both (sub)word and sentence-level tokenisers, the former for MLM and the latter for the NSP and SOP objectives. These are listed in Table~\ref{tab:tokenization}.

\begin{table}[H]
    \centering
    \small 
    \resizebox{\columnwidth}{!}{ 
    \begin{tabular}{p{0.07\textwidth} p{0.12\textwidth} p{0.12\textwidth}}
        \hline
        \textbf{Language} & \textbf{(Sub)word tokeniser} & \textbf{Sentence tokeniser} \\ \hline
        Vietnamese     & PhoBERT \citep{Nguyen2023PhoGPT}  & Underthesea \citep{underthesea} \\ \hline
        Polish        & HerBERT \citep{Mroczkowski2021HerBERT}      & NLTK for Polish \citep{bird2009natural} \\ \hline
        Georgian      & Georgian DistilBERT \citep{Davit6174_Georgian_DistilBERT_MLM} & NLTK \citep{bird2009natural} with custom adjustments to account for Georgian-specific abbreviations \\ \hline
    \end{tabular}
    }
    \caption{Language-specific (sub)word and sentence-level tokenisers used in this study.}
    \label{tab:tokenization}
\end{table}

The following table shows size (in MB) of the dataset for each language after splitting into sentences, as well as the number of tokens contained in each:

\begin{table}[H]
    \centering
    \small 
    \resizebox{\columnwidth}{!}
    { 
    \begin{tabular}{p{0.08\textwidth} p{0.03\textwidth} p{0.10\textwidth} p{0.06\textwidth}}
        \hline
        \textbf{Language} & \textbf{Size (MB)} & \textbf{No. of tokens} & \textbf{Vocab size} \\ \hline
        Vietnamese     & 477  & 964,867,879 & 19,813\\ \hline
        Polish        & 433      & 942,769,411 & 49,100\\ \hline
        Georgian      & 693 & 1,750,440,090 & 29,217\\ \hline
    \end{tabular}
    }
    \caption{Size (in MB) and no. of tokens in each dataset used for training our monolingual BERT models.}
    \label{tab:counts}
\end{table}

In addition, we calculated two mean type-to-token ratios (TTRs) for each dataset. The first involved sampling 1000 sentences and computing the average TTR for each, and the second 1000 sentence pairs. 'Token' here refers to strings tokenised by the (sub)word tokenisers listed in Table \ref{tab:tokenization}:
\begin{table}[H]
    \centering
    \small 
    \resizebox{\columnwidth}{!}{ 
    \begin{tabular}{p{0.07\textwidth} p{0.12\textwidth} p{0.12\textwidth}}
        \hline
        \textbf{Language} & \textbf{Mean sentence TTR (IQR)} & \textbf{Mean sentence-pair TTR (IQR)} \\ \hline
        Vietnamese     & 0.93337 (0.10256)  & 0.83542 (0.10256) \\ \hline
        Polish        & 0.93076 (0.11765)      & 0.86046 (0.10825) \\ \hline
        Georgian      & 0.93163 (0.10812) & 0.85734 (0.10811) \\ \hline
    \end{tabular}
    }
    \caption{Mean per-sentence and per-sentence-pair type-to-token ratios for each dataset, alongside their interquartile ranges (IQR).}
    \label{tab:counts}
\end{table}

We note that the mean sentence TTRs --- simulating the input for the MLM objective --- and mean sentence-pair TTRs --- which relate to NSP and SOP --- are very similar for all three languages. This does not therefore provide a satisfactory explanation for why Polish and Georgian models responded differently to increased depth and regularisation from Vietnamese. Computing TTRs over numbers of samples closer to the batch sizes used may provide a clearer picture of how differing TTRs contributed to performance differences.

\section{Training Details}
Each model was trained on a single NVIDIA H100 GPU for 1 epoch. We used the following settings:

\textbf{Model architecture}. All models had 8 attention heads per layer with a hidden dimension of 512.

\textbf{Optimiser}. AdamW with these parameters:
\begin{itemize}
    \item Initial learning rate: \(\ 1 \times 10^{-6}\)
    \item \(\beta_1 = 0.9, \beta_2 = 0.98\)
    \item \(\epsilon = 1 \times 10^{-12}\)
    \item Weight decay: 0.01
\end{itemize}

\textbf{Learning rate scheduler}. A cosine annealing learning rate schedule was applied, with:
\begin{itemize}
    \item Warm-up ratio: 10\% of training steps
    \item Maximum learning rate: \(\ 1 \times 10^{-4}\)
    \item Minimum learning rate: \(\ 1 \times 10^{-6}\)
\end{itemize}
The learning rate was increased linearly during the warm-up phase and subsequently followed a cosine decay schedule until training completion. All models completed training within 6 hours.

\section{Evaluation Tasks}
\label{sec:appendix_results}
\begin{table*}
  \centering
  \small 
  \begin{tabular}{p{0.15\textwidth} p{0.30\textwidth} p{0.15\textwidth} p{0.15\textwidth}}
    \hline
    \textbf{Task} & \textbf{Context supplied} & \textbf{Correct option} & \textbf{Incorrect option} \\
    \hline
    Real-world knowledge & A mammal that has wings is called a \texttt{<MASK>}. & bat & laughter \\
    \hline
    Textual entailment & The sky was clear and blue this morning. Therefore, & it did not rain this morning. & it rained this morning. \\
    \hline
    Winograd Schema Challenge & Jane knocked on Susan's door but she did not answer. \texttt{<MASK>} did not answer. & Susan & Jane \\
    \hline
  \end{tabular}
  \caption{Examples (in English) of prompts for our three additional tasks. The real-world knowledge and Winograd Schema Challenge tasks require the model to replace the \texttt{<MASK>} token(s) representing the missing word with either option, which in the languages studied could each comprise multiple tokens. English names in the latter task were all replaced by more native-sounding equivalents in the languages studied.}
  \label{tab:tasks_comparison}
\end{table*}

\begin{table*}
\centering
\small
{ 
\begin{tabular}{lrrr|rrr|rrr}
\hline
MLM+NSP & \multicolumn{3}{c}{\textbf{Vietnamese}} & \multicolumn{3}{c}{\textbf{Polish}} &  \multicolumn{3}{c}{\textbf{Georgian}} \\
Depth $\backslash$  Dropout &
\multicolumn{1}{c}{0 dr} & \multicolumn{1}{c}{0.1 dr} & \multicolumn{1}{c}{Diff.} &
\multicolumn{1}{c}{0 dr} & \multicolumn{1}{c}{0.1 dr} & \multicolumn{1}{c}{Diff} &
\multicolumn{1}{c}{0 dr} & \multicolumn{1}{c}{0.1 dr} & \multicolumn{1}{c}{Diff} \\
\hline
\multicolumn{10}{c}{Faithfulness detection}  \\
\hline
12 layers & 62.5 & 60 & -2.5 & \textbf{100} & \textbf{100} & 0 & 52.5 & 45 & -7.5 \\
8 layers  &  73.75 & 82.5 &   8.75 & \textbf{100} & \textbf{100} & 0 & 60 & 67.5 & 7.5 \\
4 layers  & \textbf{95} & 93.75 &  -1.25 & \textbf{100} & \textbf{100} & 0 & 62.5 & \textbf{71.25} & 8.75 \\
\hline  
\multicolumn{10}{c}{Real-world knowledge}  \\
\hline
12 layers & 66.67 & 70 & 3.33 & 48.89 & 51.11 & 2.22 & 66.67 & 67.82 & 1.15 \\
8 layers & \textbf{73.33} & 65.56 & -7.77 & 46.67 & 47.78 & 1.11 & 66.67 &\textbf{74.71} & 8.04 \\
4 layers & 62.22 & 65.56 & 3.34 & \textbf{55.56} & 53.33 & -2.23 & 67.82 & 58.62 & -9.2 \\
\hline  
\multicolumn{10}{c}{Textual entailment}  \\
\hline
12 layers & 40 &\textbf{63.33} & 23.33 & 76.67 & \textbf{100} & 23.33 & 46.67 & 40 & -6.67 \\
8 layers & 46.67 & 50 & 3.33 & 70 & 80 & 10 & \textbf{50} & 46.67 & -3.33 \\
4 layers & 50 & 40 & -10 & 83.33 & 93.33 & 10 & 40 & \textbf{50} & 10 \\
\hline  
\multicolumn{10}{c}{Winograd Schema Challenge}  \\
\hline
12 layers & 48.57 & 31.43 & -17.14 & 54.29 & 48.57 & -5.72 & \textbf{57.14} & \textbf{57.14} & 0 \\
8 layers & 60 & 48.57 & -11.43 & 48.57 & \textbf{57.14} & 8.57 & 54.29 & \textbf{57.14} & 2.85 \\
4 layers & \textbf{65.71} & 57.14 & -8.57 & 45.71 & 42.86 & -2.86 & 45.71 & 48.57 & 2.86 \\
\hline
\end{tabular}
}
\caption{{
Accuracy by task and language for all BERT model variants trained using MLM+NSP, , batch size 128.
(`0 dr' --- \textit{no dropout applied}, `0.1 dr' --- \textit{0.1 dropout applied}, `Diff' --- \textit{difference between both dropout settings})}}
\label{tab:mlm_nsp}
\end{table*}

\begin{table*}
\centering
\small
{ 
\begin{tabular}{lrrr|rrr|rrr}
\hline
MLM+SOP & \multicolumn{3}{c}{\textbf{Vietnamese}} & \multicolumn{3}{c}{\textbf{Polish}} &  \multicolumn{3}{c}{\textbf{Georgian}} \\
Depth $\backslash$ Dropout &
\multicolumn{1}{c}{0 dr} & \multicolumn{1}{c}{0.1 dr} & \multicolumn{1}{c}{Diff} &
\multicolumn{1}{c}{0 dr} & \multicolumn{1}{c}{0.1 dr} & \multicolumn{1}{c}{Diff} &
\multicolumn{1}{c}{0 dr} & \multicolumn{1}{c}{0.1 dr} & \multicolumn{1}{c}{Diff} \\
\hline
\multicolumn{10}{c}{Faithfulness detection}  \\
\hline
12 layers & 48.75 & 53.75 & 5.00 &  \textbf{100} & 67.5 & -32.5 & 33.75 & 43.75 & 10 \\
8 layers & 55 & 25 & -30 & 80 & \textbf{100} & 20 & 35 & 21.25 & -13.75 \\
4 layers & \textbf{65} & 61.25 & -3.75 & 85 & 77.5 & -7.5 & \textbf{56.25} & 61.25 & 5 \\
\hline  
\multicolumn{10}{c}{Real-world knowledge}  \\
\hline
12 layers & 66.67 & 66.67 & 0 & \textbf{56.67} & 55.56 & -1.11 & \textbf{68.97} & 64.37 & -4.6 \\
8 layers & 62.22 & 65.56 & 3.34 & 53.33 & 52.22 & -1.11 & \textbf{68.97} & 67.82 & -1.15 \\
4 layers & 63.33 & \textbf{71.11} & 7.78 & 50 & 45.56 & -4.44 & 63.22 & 67.82 & 4.6 \\
\hline  
\multicolumn{10}{c}{Textual entailment}  \\
\hline
12 layers & 36.67 & 46.67 & 10 & 66.67 & 73.33 & 6.66 & 50 & 45 & -5 \\
8 layers & 43.33 & 46.67 & 3.34 & 80 & \textbf{86.67} & 6.67 & 50 & \textbf{67.5} & 17.5 \\
4 layers & 43.33 & \textbf{60} & 16.67 & 73.33 & 30 & -43.33 & 40 & 46.67 & 6.67 \\
\hline  
\multicolumn{10}{c}{Winograd Schema Challenge}  \\
\hline
12 layers & 54.29 & 62.86 & 8.57 & 45.71 & 45.71 & 0 & 45.71 & 45.71 & 0 \\
8 layers & \textbf{71.43} & 57.14 & -14.29 & \textbf{48.57} & 45.71 & -2.86 & 51.43 & \textbf{62.86} & 11.43 \\
4 layers & 48.57 & 51.43 & 2.86 & \textbf{48.57} & \textbf{48.57} & 0 & 48.57 & 57.14 & 8.57 \\
\hline
\end{tabular}
}
\caption{Accuracy by task and language for all BERT model variants trained using MLM+SOP, batch size 128.}
\label{tab:mlm_sop}
\end{table*}

\begin{table*}
\centering
\small
{ 
\begin{tabular}{lrr|rr|rr}
\hline
MLM+NSP+ & \multicolumn{2}{c}{\textbf{Vietnamese}} & \multicolumn{2}{c}{\textbf{Polish}} &  \multicolumn{2}{c}{\textbf{Georgian}} \\
Batch size & 0 dr & 0.1 dr & 0 dr & 0.1 dr & 0 dr & 0.1 dr \\
\hline
\multicolumn{7}{c}{Faithfulness detection}  \\
\hline
128 & \textbf{95} & 93.75 & \textbf{100} & \textbf{100} & 62.5 & 71.25 \\
256 & 88.75 & 88.75 & \textbf{100} & \textbf{100} & 72.5 & \textbf{76.25} \\
\hline  
\multicolumn{7}{c}{Real-world knowledge}  \\
\hline
128 & 62.22 & 65.56 & \textbf{55.56} & 53.33 & 58.62 & \textbf{67.82} \\
256 & \textbf{66.67} & \textbf{66.67} & 46.67 & 44.44 & 65.52 & \textbf{67.82} \\
\hline  
\multicolumn{7}{c}{Textual entailment}  \\
\hline
128 & 50 & 40 & 83.33 & 93.33 & 40 & 56.67 \\
256 & 43.33 & \textbf{53.33} & \textbf{96.67} & 86.67 & 56.67 & \textbf{67.5} \\
\hline  
\multicolumn{7}{c}{Winograd Schema Challenge}  \\
\hline
128 & \textbf{65.71} & 57.14 & 48.57 & 45.71 & 45.71 & 48.57 \\
256 & 34.29 & 17.14 & \textbf{57.14} & 40 & 42.86 & \textbf{54.29} \\
\hline
\end{tabular}
}
\caption{Batch size comparison for MLM+NSP across Vietnamese, Polish, and Georgian tasks.}
\label{tab:mlm_nsp_batchsize}
\end{table*}

\begin{table*}
\centering
\small{ 
\begin{tabular}{llcccc}
\hline
\textbf{Model} & \textbf{Depth} & \textbf{Vietnamese} & \textbf{Polish} & \textbf{Georgian} & \textbf{Average} \\
\hline
\multirow{3}{2cm}{MLM+NSP + \\ 0 dropout}
 & 12 layers & 54.44 & 69.96 & 55.75 & 60.05 \\
 & 8 layers & 63.44 & 66.31 & 57.74 & 62.50 \\
 & 4 layers & 68.23 & 71.87 & 54.01 & 64.70 \\
\hline
\multirow{3}{2cm}{MLM+NSP + \\ 0.1 dropout}
 & 12 layers & 56.19 & 74.92 & 52.49 & 61.2 \\
 & 8 layers & 61.66 & 71.23 & 61.51 & 64.80 \\
 & 4 layers & 64.11 & 73.10 & 57.11 & 64.77 \\
\hline
\multirow{3}{2cm}{MLM+SOP + \\ 0 dropout}
 & 12 layers & 51.60 & 67.26 & 49.61 & 56.16 \\
 & 8 layers & 58.00 & 65.48 & 51.35 & 58.27 \\
 & 4 layers & 55.06 & 64.23 & 52.01 & 57.10 \\
\hline
\multirow{3}{2cm}{MLM+SOP + \\ 0.1 dropout}
 & 12 layers & 57.49 & 60.53 & 49.71 & 55.91 \\
 & 8 layers & 48.59 & 71.15 & 54.86 & 58.95 \\
 & 4 layers & 60.95 & 50.41 & 58.22 & 57.58 \\
\hline
\end{tabular}
}
\caption{Mean model performance across tasks by language, batch size 128.}
  \label{tab:mean_perf}
\end{table*}

\begin{table*}
\centering
\small{ 
\begin{tabular}{lcccc}
\hline
\textbf{Model} & \textbf{Batch size} & \textbf{Vietnamese} & \textbf{Polish} & \textbf{Georgian} \\
\hline
MLM+NSP + 0 dropout & 128 & 68.23 & 71.87 & 51.71 \\
                   & 256 & 56.59 & 75.12 & 59.39 \\
MLM+NSP + 0.1 dropout & 128 & 64.11 & 73.09 & 59.41 \\
                      & 256 & 56.47 & 73.09 & 62.09 \\
\hline
\end{tabular}
}
\caption{Mean model performance across tasks by language, with differing batch sizes and dropout levels in the MLM+NSP | 4 layers setting.}
  \label{tab:mlm_nsp_batchsize_new}
\end{table*}

\begin{table*}
  \centering
  \small 
  \begin{tabular}{p{0.15\textwidth} p{0.23\textwidth} p{0.23\textwidth} p{0.23\textwidth}}
    \hline
    \textbf{Task} & \textbf{Vietnamese} & \textbf{Polish} & \textbf{Georgian} \\
    \hline
    Faithfulness detection & MLM+NSP | 0 dropout | 4 layers | batch size 128 & all MLM+NSP models, MLM+SOP | 0 dropout | 12 layers and MLM+SOP | 0.1 dropout | 8 layers (batch sizes 128, 256 for all) & MLM+NSP | 0.1 dropout | 4 layers | batch size 256 \\
    \hline
    Real-world knowledge & MLM+NSP | 0 dropout | 8 layers | batch size 128 & MLM+SOP | 0 dropout | 12 layers | batch size 128 & MLM+NSP | 0.1 dropout | 8 layers | batch size 128 \\
    \hline
    Textual Entailment & MLM+NSP | 0.1 dropout | 12 layers | batch size 128 & MLM+NSP | 0.1 dropout | 12 layers | batch size 128; MLM+NSP | 0 dropout | 8 layers | batch size 256; MLM+NSP | 0.1 dropout | 8 layers | batch size 256 & MLM+SOP | 0.1 dropout | 8 layers | batch size 128 \\
    \hline
    Winograd Schema Challenge & MLM+SOP | 0 dropout | 8 layers | batch size 128 & MLM+NSP | 0.1 dropout | 8 layers | batch size 128; MLM+NSP | 0 dropout | 4 layers | batch size 256 & MLM+SOP | 0.1 dropout | 8 layers | batch size 128 \\
    \hline
  \end{tabular}
  \caption{Optimal hyperparameters by task for each language.}
  \label{tab:language_models}
\end{table*}

\end{document}